\documentclass[sigconf]{acmart}

\AtBeginDocument{
  }

\begin{document}

\title{GIFGuard: Proactive Forensics against Deepfakes in Facial GIFs via Spatiotemporal Watermarking}

\author{Shupeng Che}
\email{cheshuepeng@stu.xju.edu.cn}
\affiliation{
  \institution{School of Computer Science and Technology, Xinjiang University}
  \city{Urumqi}
  \country{China}
}

\author{Zhiqing Guo}
\authornote{Corresponding author.}
\email{guozhiqing@xju.edu.cn}
\affiliation{
  \institution{School of Computer Science and Technology, Xinjiang University}
  \city{Urumqi}
  \country{China}
}

\author{Changtao Miao}
\email{miaochangtao.mct@antgroup.com}
\affiliation{
  \institution{Ant Digital Technologies, Ant Group}
  \city{Hangzhou}
  \country{China}
}

\author{Dan Ma}
\email{madan@xju.edu.cn}
\affiliation{
  \institution{School of Computer Science and Technology, Xinjiang University}
  \city{Urumqi}
  \country{China}
}

\author{Gaobo Yang}
\email{yanggaobo@hnu.edu.cn}
\affiliation{
  \institution{College of Computer Science and Electronic Engineering, Hunan University}
  \city{Changsha}
  \country{China}
}

\settopmatter{printacmref=false} 
\renewcommand\footnotetextcopyrightpermission[1]{} 
\pagestyle{plain} 

\begin{abstract}
  The rapid evolution of deepfake technology poses an unprecedented threat to the authenticity of Graphics Interchange Format (GIF) imagery, which serves as a representative of short-loop temporal media in social networks. However, existing proactive forensics works are designed for static images, which limits their applicability to animated GIFs. To bridge this gap, we propose GIFGuard, the first spatiotemporal watermarking framework tailored for deepfake proactive forensics in GIFs. In the embedding stage, we propose the Spatiotemporal Adaptive Residual Encoder (STARE) to ensure robustness against high-level semantic tampering. It employs a 3D convolutional backbone with adaptive channel recalibration to capture globally coherent temporal dependencies. In the extraction stage, we design the Deep Integrity Restoration Decoder (DIRD). It utilizes a spatiotemporal hourglass architecture equipped with 3D attention to restore latent features, allowing for the accurate extraction of watermark signals even under severe facial manipulation. Furthermore, we construct GIFfaces, the first large-scale benchmark dataset curated for GIF proactive forensics to facilitate research in this domain. Extensive results show that GIFGuard achieves high-fidelity visual quality and remarkable robustness performance against deepfakes. The related code and dataset are available at https://github.com/vpsg-research/GIFGuard.
\end{abstract}

\maketitle
\begin{figure}[!t]
  \centering
  \includegraphics[width=\linewidth]{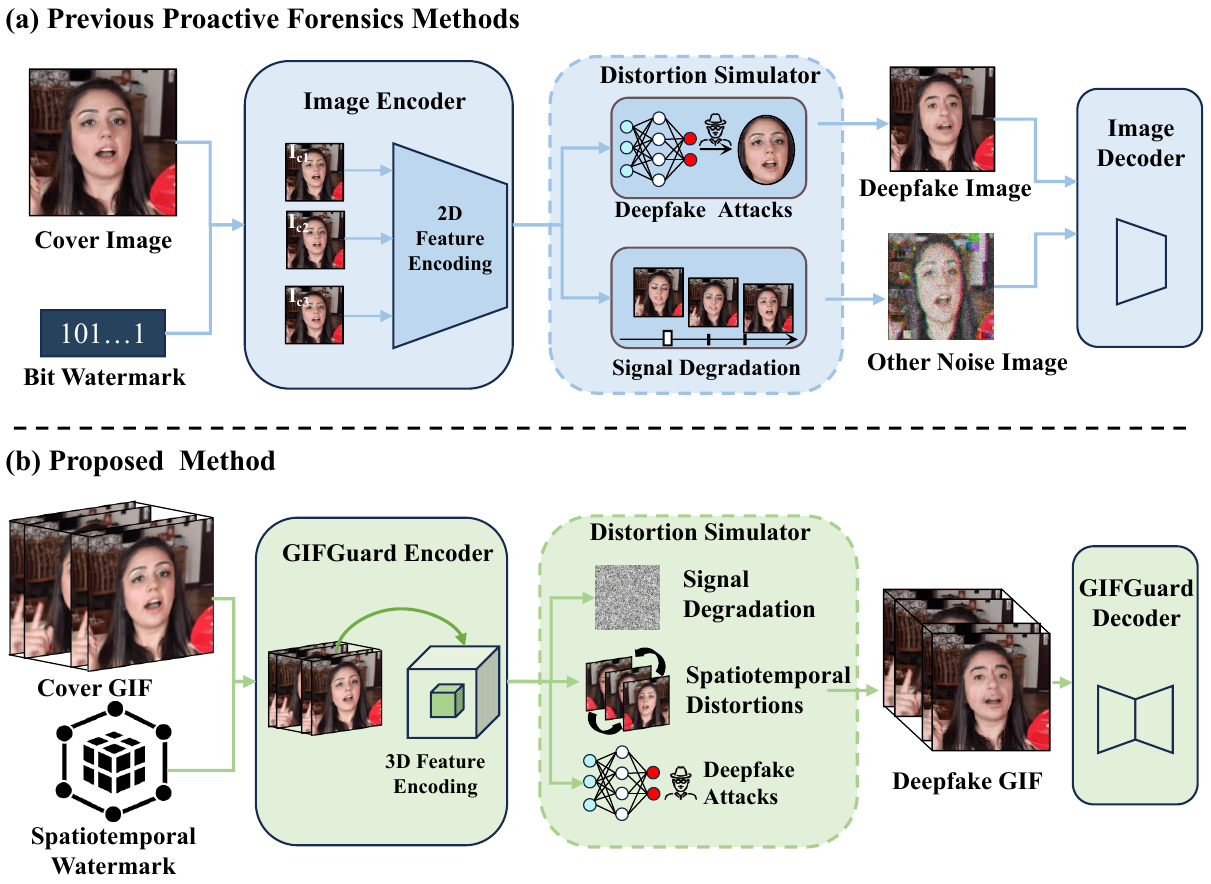}
  \caption{Comparison of proactive forensics paradigms. (a) Existing methods rely solely on spatial features, neglecting temporal correlations and failing against GIF-based distortions. (b) Our GIFGuard exploits spatiotemporal dependencies to ensure robustness against deepfakes and temporal manipulations.}
  \Description{A two-part pipeline diagram. The first part shows prior image-level proactive forensics processing GIF frames independently, which leads to inconsistent watermark recovery after deepfake or temporal distortions. The second part shows GIFGuard embedding and recovering watermark information across the full spatiotemporal GIF sequence, preserving temporal consistency under attacks.}\label{fig_1}
\end{figure}


%%
%% This command processes the author and affiliation and title formation and builds the first part of the formatted document.

\section{Introduction}
As one of the most widely circulated media in contemporary social networks, the Graphics Interchange Format (GIF) has become a fundamental medium for visual communication. Positioned between static images and videos, GIFs convey richer dynamic content than static images while maintaining smaller file sizes and more efficient transmission capabilities compared to videos. Concretely, face-centric GIFs (e.g., social emoticons and reaction clips) are extensively utilized due to their ability to capture subtle facial expressions and emotional cues~\cite{kopelman2026look}. However, the widespread usage and editable nature of GIFs, coupled with the absence of effective mechanisms for provenance tracking and content authentication, render facial content highly prone to malicious tampering. This vulnerability poses significant risks to public trust in digital media~\cite{amerini2025deepfake}. Consequently, establishing a proactive forensics mechanism to verify the authenticity and provenance of GIFs has become a critical research challenge. 

Existing proactive forensics methods enable content tracing, manipulation localization, and authenticity verification for images by embedding imperceptible and robust signals~\cite{nguyen2025survey}. However, current research has predominantly focused on spatial-domain watermarking tailored for static imagery. Representative frameworks, such as SepMark~\cite{wu2023sepmark}, MEA~\cite{jia2025uncovering}, and WaveGuard~\cite{he2025waveguard} have achieved remarkable success in protecting single images. Yet, these approaches are fundamentally incompatible with the inherent temporal structure of GIFs. As illustrated in the upper part of Figure~\ref{fig_1}, when extended to dynamic sequences, existing approaches are restricted to single image embedding strategies that treat the GIF sequence strictly as a series of discrete frames. Such frame-independent processing neglects the intrinsic temporal correlations within GIFs, causing the watermark signals to lack temporal consistency. The resulting incoherence not only impairs visual stability but also renders the defense ineffective against frame-level manipulations. 

To bridge this gap, we think about the face forensics of GIF from two perspectives: spatiotemporal consistency and semantic robustness. First, it is imperative to overcome the limitations of frame-independent paradigms prevalent in prior works. Instead of treating GIFs as a discrete set of frames, the watermarking mechanism should perceive the sequence as a unified spatiotemporal volume, leveraging global dependencies to guarantee temporal consistency. Second, regarding robustness against deepfake attacks, conventional watermarks predominantly rely on fragile high-frequency components or pixel-domain redundancies, which are inevitably purged during the generative reconstruction of faces. Therefore, the watermark should be embedded into semantically robust spatiotemporal regions, ensuring robustness against malicious tampering.

Based on these considerations, we construct a spatiotemporal modulation module by jointly training the \textbf{S}patio\textbf{t}emporal \textbf{A}daptive \textbf{R}esidual \textbf{E}ncoder (STARE) and the \textbf{D}eep \textbf{I}ntegrity \textbf{R}estoration \textbf{D}ecoder (DIRD), thereby ensuring the temporal consistency of the embedded watermark. Furthermore, we introduce a \textbf{R}ealistic \textbf{D}istortion \textbf{S}imulator (RDS) for adversarial training, which compels the model to embed watermarks into semantically robust spatiotemporal regions, thereby guaranteeing watermark robustness.

Our main contributions are summarized as follows:
\begin{itemize}
\item We propose GIFGuard, the first proactive forensic framework tailored for GIFs. It effectively bridges the gap in current forensics, breaking the frame-independent constraints of static methods and the forensic inadequacy of traditional watermarking against deepfakes.
\item We design a spatiotemporal modulation module with adversarial constraints. By integrating encoder-decoder feature modulation with adversarial learning, it embeds and restores watermarks in deep spatiotemporal representations, ensuring robustness against deepfake reconstruction.
\item We construct GIFfaces, the first large-scale GIF face forensics dataset comprising 50,000 samples, bridging the data scarcity in this domain. Experiments demonstrate that GIFGuard outperforms existing methods in visual imperceptibility and robustness, notably maintaining an extremely low bit error rate (BER $\approx$ 0.01\%) under deepfake attacks.
\end{itemize}

\section{RELATED WORK}

\subsection{Robust Watermarking in Visual Media }
\textbf{Static Image Watermarking.} Deep learning-based watermarking was initiated by HiDDeN~\cite{zhu2018hidden}, followed by numerous studies that aimed to enhance robustness against specific distortions. Methods like MBRS~\cite{jia2021mbrs}, De-END~\cite{fang2022end}, and Peng et al.~\cite{peng2025robust} focus on resisting compressions and analog noises. Recent advancements employ reversible or invariant mechanisms to enhance robustness against geometric transformations and generative texture reconstructions~\cite{zhang2024editguard, wen2023treering, chen2025reversible}. However, lacking temporal modeling, these frame-independent methods fail to maintain inter-frame consistency in dynamic media, rendering them vulnerable to tampering.

\textbf{Temporal Media Watermarking.} Temporal consistency is explicitly modeled by video watermarking~\cite{mansour2025comprehensive}. Approaches like DVMark~\cite{luo2023dvmark}, RC-VWN~\cite{chen2024robust}, and REVmark~\cite{zhang2023novel} leverage spatiotemporal features, whereas RivaGAN~\cite{zhang2019robust} and VideoSeal~\cite{fernandez2024videoseal} incorporate attention mechanisms and efficient architectures to optimize robustness and latency. However, these methods are primarily designed for standard video coding formats such as H.264/AVC. As a pioneer in the GIF domain, GIFMarking~\cite{liao2021gifmarking} introduced 3D-CNNs for spatiotemporal modeling, devising tailored defense mechanisms against temporal attacks such as frame deletion. Nevertheless, both video and GIF watermarking remain signal-level schemes targeting low-level distortions. Lacking semantic-aware defenses, they are susceptible to erasure during deepfake reconstruction.

\subsection{Proactive Forensics for Deepfake Defense} The proliferation of deepfake threats has driven the development of proactive forensics~\cite{nguyen2025survey}. Early explorations like FakeTagger~\cite{wang2021faketagger} and FaceSigns~\cite{Neekhara2024facesigns} utilized redundant encoding or semi-fragile watermarks for identity tracking, yet they often treated forensic tasks in isolation. To address this, recent methods have attempted to unify multiple forensic objectives, such as deepfake detection, source tracing, and tampering localization, through separable, landmark-aware, or identity-aware watermark designs~\cite{wu2023sepmark, wu2026all, 11552736}. Beyond framework unification, researchers have innovated embedding mechanisms: LampMark~\cite{wang2024lampmark} and WaveGuard~\cite{he2025waveguard} exploit facial landmarks and wavelet transforms, respectively, to enhance structural robustness, while another branch of research leverages generative and diffusion priors~\cite{sun2025diffmark, fernandez2023stable, yang2024gaussian} to optimize imperceptibility. Recently, strategies have evolved toward active multidimensional defenses. For instance, AdvMark~\cite{wu2024advmark} pioneers an ``assistive'' perspective to aid downstream detectors, whereas TSDF~\cite{Zheng2026Boosting} propose a dual defense mechanism to boost the persistence of proactive defense. Meanwhile, MEA~\cite{jia2025uncovering} focus on defense robustness by mitigating destructive multi-embedding attacks.

\begin{figure*}[!t]
\centering
\includegraphics[width=\linewidth, height=8.5cm, keepaspectratio]{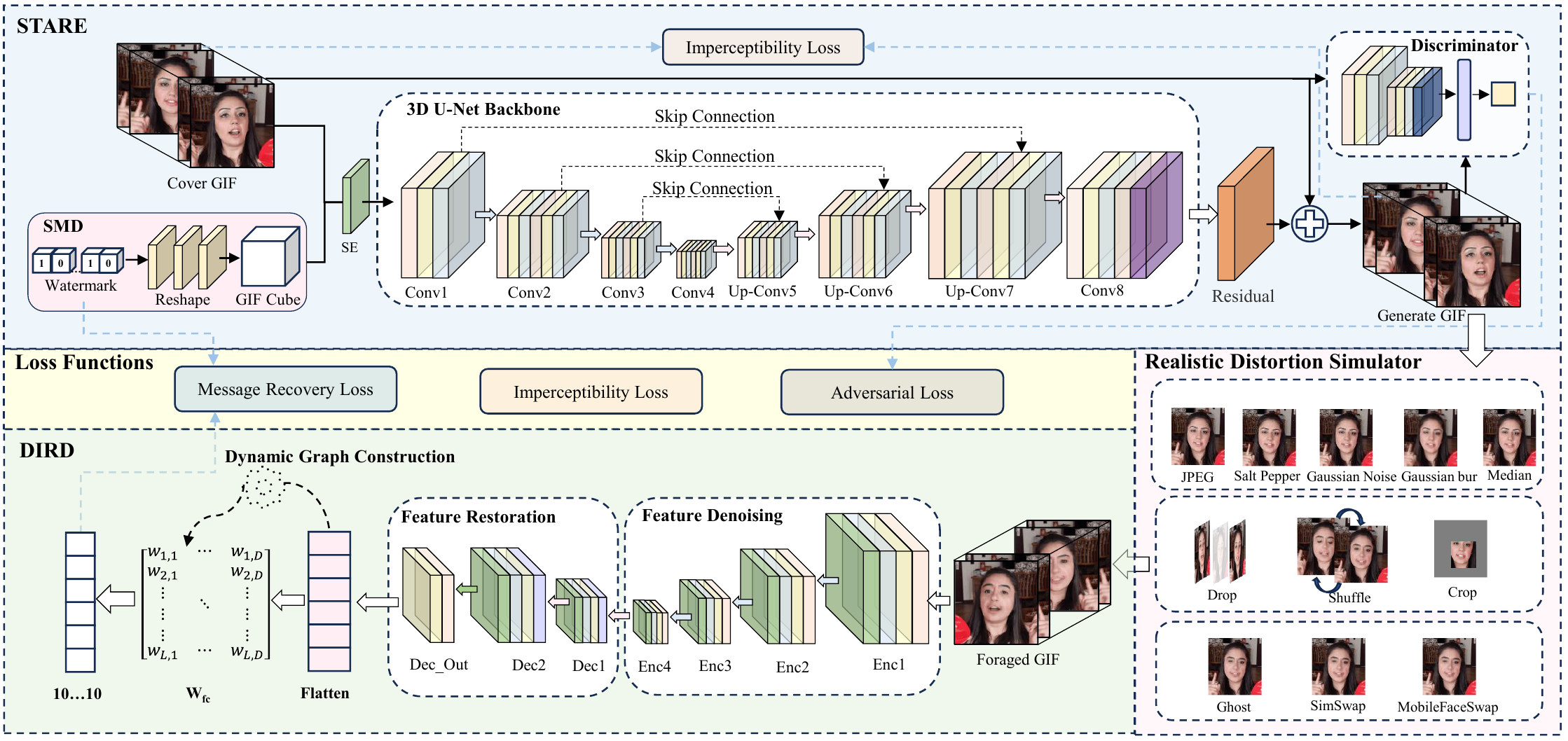}
\caption{Overview of the GIFGuard framework. The architecture consists of three key modules: (1) STARE, an encoder that embeds watermarks into the cover GIF; (2) RDS, a simulator that subjects the watermarked GIF to deepfake forgeries and various distortions; and (3) DIRD, an extractor that robustly recovers the watermark from the forged content. Additionally, a Discriminator is employed to distinguish between cover and watermarked GIFs, enforcing visual imperceptibility.}
\Description{A framework diagram showing a cover GIF and binary watermark entering the STARE encoder to produce a watermarked GIF. The watermarked GIF is passed through the Realistic Distortion Simulator, which includes deepfake attacks, geometric distortions, and signal degradation. The distorted output is then processed by the DIRD extractor to recover the watermark, while a discriminator supervises visual imperceptibility between cover and watermarked GIFs.}\label{fig_2}
\end{figure*}

Despite these advancements, existing proactive strategies are confined to the spatial domain. When applied to GIFs, they face a dual failure: (1) they cannot ensure temporal consistency, leading to visual artifacts; and (2) they lack spatiotemporal structural constraints to resist the semantic reconstruction of deepfake generators. Consequently, constructing a unified framework that achieves temporal consistency while maintaining robustness against deepfake attacks remains an unaddressed challenge.

\section{THE GIFFACES DATASET}
Existing proactive forensic studies largely rely on static face image datasets such as CelebA~\cite{liu2015deep}, CelebA-HQ~\cite{karras2017progressive}, and LFW~\cite{huang2008labeled}, which cannot adequately capture the dynamic variations and complex backgrounds of real-world GIF scenarios. To facilitate proactive face forensics in GIF scenarios, we construct GIFfaces, the first large-scale dataset tailored for proactive face GIF forensics.

\textbf{Source Data Selection.} GIFfaces is built from five mainstream video benchmarks and manually collected YouTube clips to better reflect real-world online social contexts. Specifically, we include Celeb-DF-v2~\cite{li2020celeb} to cover public-figure interviews and speeches that are susceptible to identity forgery; DFEW~\cite{10.1145/3394171.3413620} and CAER~\cite{lee2019context} to represent film and television content widely circulated on social platforms; DFDC~\cite{dolhansky2020deepfake} to provide diverse human subjects and recording conditions; and FaceForensics++~\cite{rossler2019faceforensics++} together with YouTube clips to capture visual characteristics of real-world social media videos. These complementary sources encompass diverse identities, scenes, expressions, and acquisition conditions, supporting robust generalization in practical GIF forensic scenarios.

\textbf{Data Preprocessing.} To convert video sources into GIF-specific samples, we design a unified preprocessing and annotation pipeline. We first verify video validity and perform frame-level face detection to remove corrupted or unsuitable samples. For each valid video, we extract 10 consecutive frames starting from the first frame with a detected face, and crop the face region to a resolution of $256 \times 256$.
\textbf{Format Conversion.} The cropped frame sequences are converted into the standard GIF format. During conversion, full-color frames are mapped to a palette of 256 colors, introducing irreversible color quantization. Floyd-Steinberg dithering~\cite{floyd1976adaptive} is applied to reduce banding artifacts, followed by Lempel-Ziv-Welch (LZW) lossless encoding. This process preserves the discrete color indexing and temporal dithering characteristics inherent to GIFs.
\textbf{Annotation.} Each subject is assigned a unique identity label, ensuring that the same individual maintains a consistent identifier across the dataset. This identity-consistent annotation enables principled sampling for training, validation, and testing sets, thereby enhancing the dataset's utility for deepfake forensic research. Additional details on dataset necessity, raw-data heterogeneity, and identity distribution are provided in the supplementary material.

\section{METHODOLOGY}
\subsection{Spatiotemporal Adaptive Residual Encoder}
Recognizing that deepfake manipulation mainly modifies facial semantics while preserving identity-related structures, and that GIF compression tends to degrade smooth and low-frequency regions, we design Spatiotemporal Adaptive Residual Encoder (STARE) to guide watermark embedding toward spatiotemporal regions that are both structurally stable and perceptually complex.

\textbf{Spatiotemporal Message Expansion.}
As illustrated in the top-left of Fig.~\ref{fig_2}, the original watermark $\mathbf{M} \in \{0, 1\}^L$ is a 1D binary vector, which lacks the spatiotemporal structure required to align with the GIF tensor $\mathbf{G}_{co} \in \mathbb{R}^{C \times T \times H \times W}$. To enable robustness under spatiotemporal distortions and semantic transformations, we map $\mathbf{M}$ into a distributed representation that spans the full spatiotemporal volume. Specifically, the message is projected into a compact low-resolution spatiotemporal latent volume, normalized, and then expanded via trilinear interpolation to match the resolution of $\mathbf{G}_{co}$:
\begin{align}
\mathbf{M}_{feat} = \mathcal{F}_{up}(\mathcal{R}(\mathcal{F}_{fc}(\mathbf{M})))
\end{align}
This design distributes the watermark to be redundantly encoded across both spatial and temporal dimensions, reducing reliance on any single location. As a result, even when parts of the content are altered or re-synthesized by deepfake models, sufficient information can still be preserved for reliable recovery.

\textbf{Adaptive Feature Fusion.} 
To ensure that watermark embedding to leverage both visual content and message-aware cues, we first concatenate the cover GIF feature and the expanded message feature. Specifically, the joint feature $\mathbf{X} = [\mathbf{G}_{co}; \mathbf{M}_{feat}]$ is processed by a 3D Squeeze-and-Excitation (SE) block to capture global spatiotemporal channel dependencies.
Global average pooling is applied to aggregate contextual information and produce a channel descriptor $\mathbf{z}$:
\begin{align}
z_c = \frac{1}{T \times H \times W} \sum_{t=1}^T \sum_{h=1}^H \sum_{w=1}^W \mathbf{X}_{c,t,h,w}
\end{align}
The resulting attention weights emphasize informative spatiotemporal channels while suppressing less relevant ones. This produces a more robust feature representation for residual generation during generative reconstruction.

\textbf{Hierarchical Spatiotemporal Residual Learning.} 
To align watermark embedding with semantically meaningful structures, we adopt a hierarchical spatiotemporal modeling strategy based on a 3D U-Net backbone. This design captures both global context and local variations, enabling the network to generate a spatiotemporally adaptive residual map that controls the location and strength of watermark embedding. Regions with higher structural or motion complexity receive stronger residuals, while smoother regions are minimally perturbed to preserve visual quality. The final watermarked GIF $\mathbf{G}_{w}$ is formulated as:
\begin{align}
\mathbf{G}_{w} = \mathbf{G}_{co} + \alpha \cdot \mathcal{F}_{out}(\mathbf{X}_{dec})
\end{align}
where $\mathbf{X}_{dec}$ denotes the decoded spatiotemporal feature volume produced by the 3D U-Net decoder, $\mathcal{F}_{out}$ is the final residual prediction head, and $\alpha$ controls the trade-off between visual imperceptibility and watermark robustness. This residual formulation enables controlled watermark embedding while preserving visual fidelity.

\subsection{Realistic Distortion Simulator}
To address the challenges of real-world environments, we design the Realistic Distortion Simulator (RDS), which integrates three distinct categories of distortions to simulate the complex transmission pipeline of GIFs.

\textbf{Deepfake Attacks.} Proactive robustness is achieved by integrating frozen pre-trained generative models (SimSwap~\cite{chen2020simswap}, MobileFaceSwap~\cite{xu2022mobilefaceswap}, Ghost~\cite{groshev2022ghost}) as fixed attack simulators during training. These simulators provide realistic semantic reconstruction perturbations, compelling the encoder to learn features that remain robust against the reconstruction patterns of different deepfake generators. The selection of these specific models is motivated by their robustness to complex backgrounds. In contrast, alternative methods such as StarGAN~\cite{choi2020stargan} and GANimation~\cite{pumarola2018ganimation} are primarily tailored for face-dominant datasets. Consequently, applying them to our unconstrained scenes inevitably leads to severe image distortion and corrupts the background content. \textbf{Geometric Distortion.} We simulate GIF re-composition by introducing temporal perturbations (frame drop, shuffle) and spatial random cropping. These operations disrupt the $(T, H, W)$ structure, guiding the model to learn a distributed watermark distribution independent of absolute spatiotemporal coordinates. \textbf{Signal Degradation.} Addressing the non-differentiable nature of GIF palette quantization is critical. We propose a compound approximation strategy that utilizes a differentiable JPEG layer as a robust proxy. By applying block-wise DCT transforms coupled with soft-rounding approximations on frequency coefficients, we simulate the high-frequency information loss and banding artifacts characteristic of aggressive GIF compression. Additionally, we integrate differentiable implementations of blurring and additive noise to mimic transmission degradation, allowing the major signal degradation operations to be incorporated into training in a gradient-compatible manner where applicable.

\textbf{Adaptive Learning Strategy.}
We employ an adaptive learning strategy based on RDS. Utilizing a probability climbing mechanism to optimize convergence under complex distortions, the strategy constructs an adversarial path of increasing difficulty: it begins by establishing a precision baseline under identity mapping, gradually introduces signal and temporal distortions for basic robustness, and culminates by elevating the sampling probabilities of random crop and deepfake simulators. This progression compels the model to encode watermarks into topology-invariant representations insensitive to semantic alterations, ensuring robust defense in highly adversarial environments.

\subsection{Deep Integrity Restoration Decoder}
The Deep Integrity Restoration Decoder (DIRD) is designed to recover embedded watermarks from spatiotemporally and semantically distorted GIFs. Under deepfake manipulation, watermark signals are often fragmented, redistributed, and partially suppressed due to generative reconstruction. To address this, DIRD treats watermark extraction as a spatiotemporal restoration problem, aiming to reconstruct watermarking from degraded and misaligned features. Specifically, the DIRD architecture comprises the following core components:

\textbf{Attentive Feature Denoising.}
During extraction, the input GIF contains artifacts introduced by deepfake generation and compression. To mitigate their impact, we employ 3D convolutional layers with SE-based feature modulation to suppress unreliable responses and retain informative features. By leveraging global context, the model reduces interference from distorted regions and preserves features that are more consistent with the embedded watermark.

\textbf{Spatiotemporal Feature Restoration.}
To counter signal fragmentation caused by spatiotemporal distortions, the expanding path incorporates a spatiotemporal integrity reconstruction module. It utilizes 3D transposed convolutions coupled with auxiliary SE-Blocks to progressively restore spatiotemporal resolution while refining feature coherence. This design re-aligns and reassembles fractured features, leveraging deep semantic cues to restore the structural integrity of the watermark signal prior to decoding.

\begin{figure*}
\centering
\includegraphics[width=\linewidth, height=9cm, keepaspectratio]{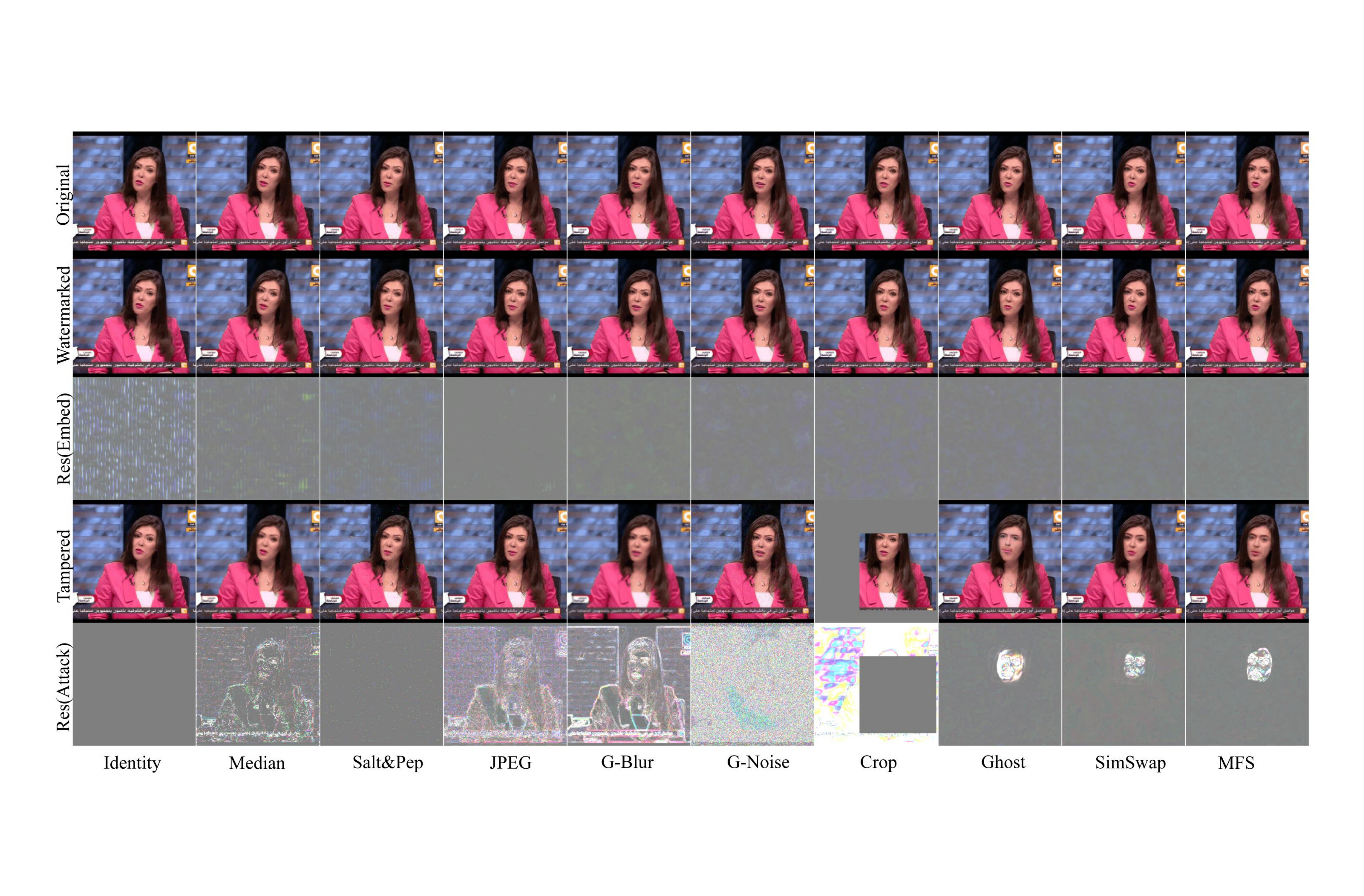}
\caption{Visualization of visual imperceptibility and robustness against mixed attacks. Rows 1--3 illustrate the high fidelity of the embedding process: the watermarked frames (Row 2) are visually indistinguishable from the originals (Row 1), as evidenced by the sparse embedding residuals (Row 3). Rows 4--5 demonstrate the diversity of the attack simulation. The tampered frames (Row 4) cover a spectrum from signal-level noises (e.g., JPEG, Crop) to semantic deepfake forgeries (e.g., SimSwap), while the attack residuals (Row 5) visualize the intensity of distortions that the framework effectively counteracts.}
\Description{A multi-row qualitative comparison of GIF frames. The upper rows compare original frames, watermarked frames, and embedding residuals, showing that watermark changes are sparse and visually subtle. The lower rows show attacked frames and attack residuals under mixed distortions, including signal degradation, cropping, and deepfake manipulation, illustrating that GIFGuard is evaluated under both low-level and semantic changes.}\label{fig_3}
\end{figure*}

\textbf{Adaptive Global Projection Head.}
After spatiotemporal feature restoration, the restored feature volume is normalized and flattened into a compact vector $\mathbf{v}_{flat}$, which is then passed through a fully connected prediction head to recover the embedded bit sequence. To streamline the deployment across different GIF specifications, we implement an automatic dimensionality alignment mechanism. Unlike static architectures with hard-coded dimensions, our design employs a structural inference strategy: during initialization, the model executes a profiling pass (using a dummy input) to deduce the exact dimensionality of the flattened feature vector $\mathbf{v}_{flat}$. This allows the final mapping matrix $\mathbf{W}_{fc}$ to be dynamically instantiated, ensuring precise alignment with the specific $(T, H, W)$ configuration without manual architectural adjustments.
\begin{equation}
\mathbf{y}_{logits} = \mathbf{W}_{fc} \cdot \mathbf{v}_{flat} + \mathbf{b}_{fc}
\end{equation}
where $\mathbf{v}_{flat}$ is the flattened vector of the restored features.

\subsection{Objective Functions}
To balance visual quality and robustness, we optimize the encoder and decoder with a multi-task objective, while training the discriminator with a separate adversarial objective. The total objective is formulated as:
\begin{equation}
\mathcal{L}_{\text{total}} = \mathcal{L}_{\text{imp}} + \lambda_{\text{adv}}\mathcal{L}_{\text{adv}} + \lambda_{\text{msg}}\mathcal{L}_{\text{msg}}
\end{equation}
where $\lambda_{\text{adv}}$ and $\lambda_{\text{msg}}$ are balancing coefficients. We employ a dynamic weighting strategy for $\lambda_{\text{msg}}$, increasing it linearly during training to emphasize bit recovery in later stages.

\textbf{Imperceptibility Loss.} We define $\mathcal{L}_{\text{imp}}$ as a weighted combination of MSE and LPIPS to preserve pixel-level fidelity and perceptual consistency:
\begin{equation}
\mathcal{L}_{\text{imp}} =\|\mathbf{G}_{co} - \mathbf{G}_w\|_2^2+ \beta \, \mathrm{LPIPS}(\mathbf{G}_{co}, \mathbf{G}_w)
\end{equation}
where LPIPS is computed with an AlexNet-based perceptual model to capture perceptual distortions often missed by pixel-wise MSE, and $\beta$ controls its contribution.

\textbf{Adversarial Loss.} To suppress artifacts and enhance realism, a discriminator $A$ is trained to distinguish watermarked GIFs from originals. The encoder optimizes the non-saturating adversarial loss to deceive $A$:
\begin{equation}
\mathcal{L}_{adv} = -\mathbb{E}_{\mathbf{G}_w}[\log(A(\mathbf{G}_w))]
\end{equation}

\begin{table}[!t]
  \caption{Quantitative comparison of visual imperceptibility. Parentheses indicate each baseline's media domain.}\label{tab_1}
    \begin{tabular*}{\linewidth}{l@{\extracolsep{\fill}}ccc}
        \toprule
        Model             &  PSNR ($\uparrow$) & SSIM ($\uparrow$) & LPIPS ($\downarrow$) \\
        \midrule
        GIFMarking (GIF)   &  40.37  &  0.9913  &  -   \\
        RivaGAN (Video)    &  42.71  &  0.9540  & -  \\
        REVmark (Video)    &  37.53  &   -    &  0.0384  \\
        VideoSeal (Video)  & 48.02 &  0.9980 & 0.0130 \\
        GIFGuard (Ours)    &  49.54  & 0.9960  &  0.0035  \\
        \bottomrule
    \end{tabular*}     
\end{table}

\textbf{Robust Message Recovery Loss.} The core defense capability is enforced by minimizing the binary cross-entropy between the original watermark $\mathbf{M}$ and the predicted bit probabilities $\hat{\mathbf{M}}$ recovered from the distorted GIF by the RDS:
\begin{equation}
\mathcal{L}_{msg} = -\frac{1}{L} \sum_{i=1}^{L} [M_i \log(\hat{M}_i) + (1 - M_i) \log(1 - \hat{M}_i)]
\end{equation}

\begin{table*}[!t]
   \caption{Quality assessment of the extracted watermark images. The retrained GIFMarking baseline maintains structural integrity under standard distortions but fails under deepfake attacks, where the VIF drops to negligible levels.}\label{tab_2}
    \resizebox{\textwidth}{!}{
        \begin{tabular}{lcccccccccccc}
            \toprule
            & \multicolumn{8}{c}{Signal Degradation \& Geometric Distortion} & \multicolumn{4}{c}{Deepfake Attacks} \\
            \cmidrule(lr){2-9} \cmidrule(lr){10-13}
            \textbf{Metric} & Identity & G-blur & Salt\&Pep & Median & JPEG & F-Repl & Drop & \textbf{Avg.} & Ghost & MFS & SimSwap & \textbf{Avg.} \\
            \midrule

            PSNR ($\uparrow$) & 36.56 & 25.92 & 35.10 & 32.10 & 36.55 & 28.54 & 20.22 & 30.71 & 7.10 & 7.10 & 7.10 & 7.10 \\
            SSIM ($\uparrow$) & 0.9831 & 0.9161 & 0.9721 & 0.9523 & 0.9831 & 0.9331 & 0.8612 & 0.9430 & 0.5281 & 0.5281 & 0.5281 & 0.5281 \\
            VIF ($\uparrow$)  & 0.8090 & 0.5922 & 0.7181 & 0.6200 & 0.8086 & 0.5978 & 0.7317 & 0.6970 & 0.0032 & 0.0032 & 0.0032 & 0.0032 \\

            \bottomrule
        \end{tabular}
    }
\end{table*}

\begin{table*}[htbp]
    \caption{Quantitative comparison of robustness (BER \%, $\downarrow$) against video watermarking baselines. Constrained by its maximum embedding capacity, RivaGAN is evaluated at a 32-bit payload, whereas our method and other baselines are tested at the rigorous 128-bit target. }\label{tab_3}
    \resizebox{\textwidth}{!}{
        \begin{tabular}{lcccccccccccccc}
            \toprule
            & \multicolumn{9}{c}{Signal Degradation \& Geometric Distortion} & \multicolumn{4}{c}{Deepfake Attacks} \\
            \cmidrule(lr){2-10} \cmidrule(lr){11-14}
            \textbf{Method} & Identity & G-blur & G-Noise & Salt\&Pep & Median & JPEG & Drop & Crop & \textbf{Avg.} & Ghost & MFS & SimSwap & \textbf{Avg.}\\
            \midrule
        
            RivaGAN & 0.0296 & 0.2292 & 0.0599 & 0.6165 & 0.0394 & 0.0469 & 24.4643 & 0.1005 & 3.1983 & 2.5362 & 2.8618 & 2.4972 & 2.6317\\
            REVmark& 21.4846 & 20.9721 & 22.9186 & 22.9962 & 21.7625 & 23.2139 & 25.1771 & 23.9351& 22.8075 & 22.7905 & 23.6268 & 22.5183 & 22.9785\\
            VideoSeal & 5.9781 & 3.9847 & 3.8287 & 9.0857 & 6.6736 & 17.8483 & 27.1436 & 11.7178 & 10.7826 & 19.2794 & 19.2670 & 19.4270 & 19.3245\\
            GIFGuard & 0.0023 & 0.0208 & 0.0326 & 0.0419 & 0.0036 & 0.0560 & 0.0167 & 0.1008 & \textbf{0.0343} & 0.0128 & 0.0076 & 0.0112 & \textbf{0.0105} \\
            
            \bottomrule
        \end{tabular}
        }  
\end{table*}

\section{Experiments}
\subsection{Experimental Settings}
\textbf{Dataset.}
We evaluate GIFGuard on the proposed GIFfaces dataset introduced in Section~3. Considering the computational cost of end-to-end 3D adversarial training, we curate a representative subset split of 10,000 training samples, 3,000 validation samples, and 3,000 testing samples. The subset is constructed by jointly considering identity separation, source domains, scene contexts, and viewpoints, so as to preserve the diversity of GIFfaces. This setup ensures experimental feasibility and provides a fair evaluation to confirm that our method generalizes well to GIFs of unseen identities.

\textbf{Implementation Details.}
Our framework is implemented using PyTorch, utilizing Automatic Mixed Precision (AMP) to optimize GPU memory efficiency. All experiments are conducted on NVIDIA A40 GPUs. The input is a 5D tensor of shape $(B, C, T, H, W)$. Specifically, we fix the temporal dimension at $T=10$, following common practice in GIF and video watermarking, and resize frames to $256 \times 256$ using a batch size of 8, while setting the watermark payload length to $L=128$ bits. The model is trained using the Adam optimizer with an initial learning rate of $1 \times 10^{-4}$.

\textbf{Distortion Simulation Setup.}
Following the Realistic Distortion Simulator (RDS), we establish a comprehensive attack environment and define the abbreviations used in the evaluation tables. Specifically, the signal degradation processes comprise Gaussian noise (G-Noise, $\sigma=0.05$), Gaussian blur (G-blur) applying a $5\times5$ kernel ($\sigma=1.0$), and salt-and-pepper noise (Salt\&Pep) with a $0.5\%$ ratio, alongside $3\times3$ 3D median filtering (Median) and JPEG compression (JPEG) with a quality factor of 75. Furthermore, spatiotemporal distortions introduce random spatial cropping (Crop) defined by a scale ratio $s \in [0.8, 1.0]$, frame dropping (Drop) with a probability of approximately 0.7, and frame replacement (F-Repl). Finally, pre-trained SimSwap, MobileFaceSwap (MFS), and Ghost are integrated as fixed non-linear operators to simulate deepfake attacks.

\textbf{Training Strategy.}
To implement the probability climbing mechanism, we train GIFGuard in four progressive stages. Epochs 1--3 use clean samples to establish a stable bit-recovery baseline before introducing distortions. Epochs 4--10 introduce signal degradation and temporal perturbations for basic robustness. Epochs 11--20 add deepfake attacks and random cropping with an initial sampling probability of $p=0.3$, enabling adaptation to semantic reconstruction and spatial misalignment. Finally, epochs 21--50 increase the sampling probability of deepfake and cropping simulators to $p=0.6$, strengthening robustness under severe attacks while maintaining stable convergence.

\subsection{Visual Quality Evaluation}
Visual imperceptibility is a prerequisite for covert watermarking. As presented in Table~\ref{tab_1}, GIFGuard demonstrates superior fidelity, establishing a new benchmark for GIF watermarking quality. Notably, our method achieves an exceptionally low LPIPS score of 0.0035, significantly surpassing the advanced VideoSeal baseline. This confirms that the introduced artifacts are virtually indistinguishable. Furthermore, in terms of pixel-level fidelity, our model reaches a PSNR of 49.54 dB, outperforming frameworks like RivaGAN. While our SSIM is merely 0.0020 lower than VideoSeal, by marginally sacrificing absolute structural similarity, we achieve superior robustness against complex spatiotemporal manipulations via rigorous adversarial training. We attribute this high imperceptibility to the Spatiotemporal Adaptive Residual Encoder (STARE), which treats the sequence as a unified volume to optimize embedding based on global spatiotemporal dependencies, rather than performing independent frame embedding. Fig.~\ref{fig_3} visually corroborates these results, where the highly sparse embedding residuals render the watermarked frames indistinguishable from the originals.

\subsection{Robustness Testing}
To comprehensively evaluate robustness, we conduct two complementary evaluations. First, we analyze GIFMarking, the only existing watermarking scheme specifically designed for animated GIFs, to diagnose the limitation of conventional GIF watermarking under deepfake reconstruction. Second, we compare GIFGuard with representative video watermarking baselines for bit-level recovery robustness. Since GIFMarking adopts image watermarks, whereas the other methods adopt bit watermarks, we report them with task-specific metrics in Table~\ref{tab_2} and Table~\ref{tab_3}, respectively. All baselines are evaluated on GIFfaces under the same applicable attack settings.

\subsubsection{Failure Analysis of the GIF Watermarking Baseline}
We select GIFMarking as the representative baseline for GIF forensics and retrain it with deepfake data to ensure a fair and reasonable evaluation. As presented in the left section of Table~\ref{tab_2}, GIFMarking exhibits excellent stability under standard distortions, maintaining high fidelity with an average SSIM of 0.9430 and VIF of 0.6970. This confirms its effectiveness in handling signal degradation and temporal operations. However, under deepfake attacks, its performance collapses across all metrics. The SSIM drops to 0.5281, and the VIF decreases to a negligible 0.0032. This contrast indicates that although the coarse structure of the extracted watermark is partially retained, the fine-grained texture information essential for reliable watermark extraction is almost completely destroyed.

\textbf{Reason for Failure.} This collapse stems from the intrinsic vulnerability of image watermarking, which relies on pixel-level residuals. Deepfake attacks do not simply perturb pixels, but rebuild facial content through semantic transformations. This process destroys the embedded watermark information. Thus, while GIFMarking is robust against conventional spatial and temporal attacks, it remains vulnerable to deepfake reconstruction.

\subsubsection{Comparison with Video Watermarking Baselines}
Given the scarcity of multi-bit watermarking schemes tailored for GIFs, we benchmark our method against representative video watermarking frameworks: RivaGAN, REVmark, and VideoSeal. This selection is motivated by the temporal-sequence nature shared by GIFs and videos, which makes video watermarking methods relevant baselines for evaluating temporal-feature robustness. We employ bit error rate (BER) as the primary quantitative metric.

\textbf{Performance under Signal Degradation and Geometric Distortion.} As detailed in Table~\ref{tab_3}, existing video baselines exhibit clear limitations under the target payload of 128 bits. REVmark and VideoSeal suffer substantial performance degradation at this capacity, with average BERs of 22.81\% and 10.78\%, respectively. RivaGAN is inherently limited to a 32-bit payload and therefore cannot support the target 128-bit capacity. With a substantially larger payload of 128 bits, GIFGuard maintains high visual quality and achieves an average BER of only 0.0343\%, demonstrating a better trade-off between payload capacity and imperceptibility. Furthermore, existing baselines remain vulnerable to temporal desynchronization; for example, RivaGAN's BER increases to 24.46\% under frame dropping. By modeling GIF sequences as a unified spatiotemporal volume, GIFGuard leverages global correlations to ensure reliable watermark extraction against both temporal desynchronization and geometric distortions.

\textbf{Robustness under Deepfake Attacks.} The right section of Table~\ref{tab_3} highlights a fundamental divergence in deepfake resistance. Traditional video methods, primarily optimized for robustness against conventional signal distortions, fail to withstand non-linear semantic tampering. REVmark and VideoSeal produce high BERs of 22.98\% and 19.32\%, respectively, suggesting that facial reconstruction severely disrupts their watermark signals. While RivaGAN presents a seemingly moderate average BER of 2.63\%, this performance is merely a consequence of its strictly constrained 32-bit payload capacity. Even under this relaxed condition, its error rate remains over two orders of magnitude higher than our method. Conversely, GIFGuard demonstrates exceptional robustness to deepfake attacks, suppressing the average BER to a negligible 0.0105\%. This superiority stems from our proactive defense strategy: by integrating the proposed realistic distortion simulator into the adversarial training loop, GIFGuard compels the encoder to shift embedding focus from fragile shallow textures to semantically robust spatiotemporal regions. This ensures the watermark remains retrievable even after the generative model completely reconstructs the facial appearance.

\begin{table*}[htbp]
     \caption{Ablation study of the 3D SE module under different attack types. The results show that removing the module leads to a significant increase in bit error rate (BER\%, $\downarrow$).}\label{tab_4}
    \resizebox{\linewidth}{!}{%
     \begin{tabular}{lccccccccccccc} 
         \toprule
         & \multicolumn{9}{c}{\textbf{Signal Degradation \& Geometric Distortion}} & \multicolumn{4}{c}{\textbf{Deepfake Attacks}} \\
         \cmidrule(lr){2-10} \cmidrule(lr){11-14}
        \textbf{Method} & Identity & G-blur & G-Noise & Salt\&Pep & Median & JPEG & Drop & Crop & \textbf{Avg.} & Ghost & MFS & SimSwap & \textbf{Avg.} \\ 
         \midrule
         With Module & 0.0023 & 0.0208 & 0.0326 & 0.0419 & 0.0036 & 0.0560 & 0.0167 & 0.1008 & \textbf{0.0343} & 0.0128 & 0.0076 & 0.0112 & \textbf{0.0105} \\
         w/o Module & 0.0036 & 0.1344 & 0.0974 & 0.2161 & 0.0036 & 0.1339 & 0.5432 & 0.0307 & \textbf{0.1454} & 0.0229  &  0.0188   &  0.0302   & \textbf{0.0240} \\ 
         \bottomrule
    \end{tabular}%
     }
 \end{table*}

 \begin{table*}[!t] 
  \caption{Ablation study of decoding head architectures on model robustness (BER \%).}\label{tab_5}  
  \centering
    \resizebox{\linewidth}{!}{%
    \begin{tabular}{lccccccccccccc}
         \toprule
         & \multicolumn{9}{c}{\textbf{Signal Degradation \& Geometric Distortion}} & \multicolumn{4}{c}{\textbf{Deepfake Attacks}} \\
         \cmidrule(lr){2-10} \cmidrule(lr){11-14}
        \textbf{Module} & Identity & G-blur & G-Noise & Salt\&Pep & Median & JPEG & Drop & Crop & \textbf{Avg.} & Ghost & MFS & SimSwap & \textbf{Avg.} \\ 
         \midrule
          Global Pooling    & 49.55 & 49.83 & 49.79 & 49.90 & 49.87 & 49.94 & 49.92 & 49.97 & \textbf{49.85} & 49.91  &  49.97   &  50.05   & \textbf{49.98} \\  
          Grid Interpolation  & 0.26 & 2.75 & 3.28 & 9.32 & 0.35 & 3.92 & 2.67 & 12.50 & \textbf{4.38}   & 6.83  &  5.66   &  9.48   & \textbf{7.32} \\
          Ours (Full)     & 0.0023 & 0.0208 & 0.0326 & 0.0419 & 0.0036 & 0.0560 & 0.0167 & 0.1008 & \textbf{0.0343} & 0.0128 & 0.0076 & 0.0112 & \textbf{0.0105} \\
         \bottomrule
      \end{tabular}%
      }     
\end{table*}

\subsection{Generalization and Deployment Analysis}
We further conduct experiments to evaluate the generalization ability and deployment practicality of GIFGuard.

\textbf{Generalization Ability.}
First, to assess robustness against unseen deepfake algorithms, we test GIFGuard on VividFace~\cite{shao2026vividface}, a diffusion-based generator not used during training, and obtain a BER of 0.2650 without retraining or fine-tuning. Second, for variable-length GIFs, we process long sequences in fixed-length chunks and aggregate the recovered bits through majority voting. On 50-frame GIFs, this strategy achieves a BER of 0.01\% under deepfake attacks, showing that GIFGuard can be extended beyond the fixed 10-frame training setting. Third, to simulate social platform re-encoding, we evaluate format transcoding and obtain extraction accuracies of 92.10\% under MP4 compression and 95.96\% under WebM conversion. These results indicate that GIFGuard has a certain degree of generalization ability.

\textbf{Deployment Practicality.}
We also evaluate deployment efficiency, with detailed results provided in the supplementary material. Although GIFGuard has a larger parameter count mainly due to the decoder, its computational cost is 197.95 GFLOPs, comparable to GIFMarking and VideoSeal and lower than REVmark and RivaGAN. In measured inference speed, GIFGuard reaches 355.20 FPS for encoding and 382.60 FPS for decoding, indicating practical real-time inference. The main efficiency limitation lies in memory usage, which motivates future lightweight decoder designs.

\subsection{Ablation Studies}
\begin{figure}[!t]
\centering
\includegraphics[width=\linewidth]{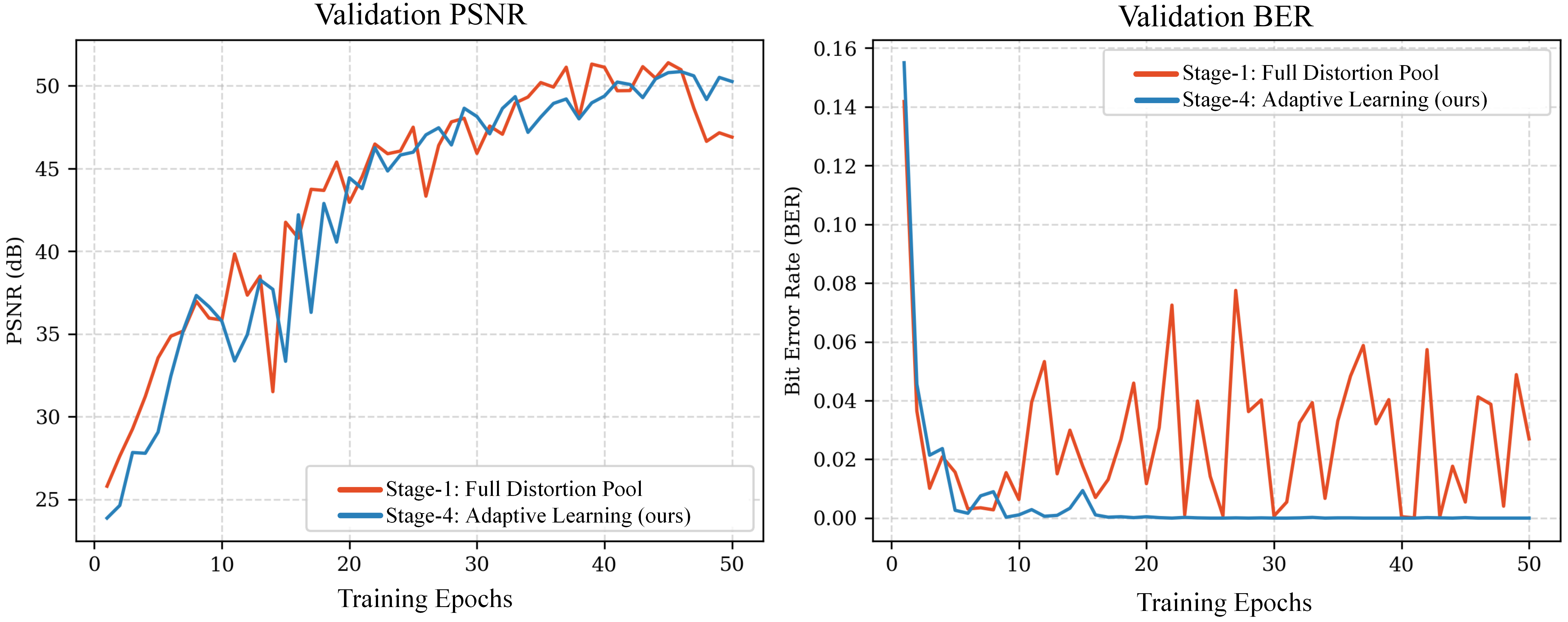}
\caption{Ablation study on the efficacy of the learning strategy. Our method achieves a higher steady-state PSNR (left) and more stable BER convergence (right).}
\Description{Two line charts comparing the proposed adaptive learning strategy with a baseline training strategy. The left chart plots PSNR over training and shows the proposed strategy converging to higher visual quality. The right chart plots bit error rate over training and shows the proposed strategy producing lower and more stable decoding error than the baseline.}\label{fig_4}
\end{figure}

\textbf{Effectiveness of Adaptive Learning Strategy.} We compare GIFGuard with a baseline trained by uniformly sampling distortions from the full distortion pool, with detailed results shown in Fig.~\ref{fig_4}. The baseline exhibits severe BER oscillations, as premature exposure to complex high-intensity distortions destabilizes the training process. In contrast, our probability climbing mechanism progressively increases attack difficulty, leading to more stable optimization. This strategy improves PSNR by approximately 3.36 dB, reaching 50.25 dB, while maintaining a near-zero BER. These results demonstrate the effectiveness of adaptive learning in stabilizing convergence and balancing robustness with visual fidelity.

\textbf{Role of Adaptive Feature Recalibration.} To verify the efficacy of the spatiotemporal Squeeze-and-Excitation (SE) attention mechanisms integrated into STARE and DIRD, we conduct an ablation study by removing them from the network. The results show clear performance degradation across different attack types. Specifically, the average BER under signal degradation and geometric distortions increases from 0.0343\% to 0.1454\%. More importantly, as shown in Table~\ref{tab_4}, the BER under deepfake attacks increases by approximately 2.3x. These results confirm that adaptive channel-wise feature recalibration helps prioritize informative spatiotemporal representations and suppress complex generative distortions.

\textbf{Ablation on Decoding Head Architectures.} We compare the proposed Adaptive Global Projection Head with global pooling and grid interpolation  (e.g., SepMark~\cite{wu2023sepmark}). As reported in Table~\ref{tab_5}, global pooling leads to decoding failure with a BER close to 50\%, since excessive feature compression removes critical watermark information. Similarly, grid interpolation compresses and smooths dynamic features by aligning them into a fixed 3D grid, causing spatiotemporal information loss and yielding a BER of 7.32\% under deepfake attacks. In contrast, our method preserves richer spatiotemporal features through adaptive global projection, thereby improving watermark recovery under semantic reconstruction and achieving a deepfake BER of 0.0105\%.

\section{Conclusion}
In this paper, we propose GIFGuard, the first spatiotemporal proactive forensic framework tailored to GIFs against deepfakes. To bridge the gap of spatiotemporal inconsistency, we construct a modulation module by jointly optimizing the STARE and DIRD. This design effectively captures global dependencies within the unified spatiotemporal volume, ensuring the embedded signals maintain temporal coherence and visual fidelity. Crucially, we introduce the Realistic Distortion Simulator for adversarial training to simulate facial reconstruction artifacts. This mechanism drives the model to embed watermarking information into semantically stable regions, guaranteeing robustness against sophisticated deepfake manipulations. Extensive experiments on the GIFfaces dataset demonstrate GIFGuard's superiority over existing baselines. Building upon this robust forensic baseline, future work will aim to deploy lightweight variants for real-time edge applications, while generalizing this proactive paradigm to broader generative video formats to maintain trackability against continuously evolving forgeries.

\begin{acks}
This work was supported in part by the National Natural Science Foundation of China under Grants 62462060 and 62302427, in part by the Central Government Guides Local Science and Technology Development Fund Projects under Grant ZYYD2026ZY21, and in part by the Doctoral Innovation Fund Project of Xinjiang University under Grant XJDX2025YJS090.
\end{acks}

\bibliographystyle{ACM-Reference-Format}
\bibliography{main}

% If your work has an appendix, this is the place to put it.

\clearpage
\appendix

\section*{Supplementary Material}

\setcounter{figure}{0}
\setcounter{table}{0}
\renewcommand{\thefigure}{S\arabic{figure}}
\renewcommand{\thetable}{S\arabic{table}}

\section{Dataset Analysis}

To complement the main text, we provide additional details on the construction and analysis of GIFfaces. This section includes comparisons with traditional static face datasets, statistics of the raw video sources, and visualizations of the preprocessing pipeline.

\subsection{Comparative Analysis with Traditional Static Face Image Datasets}
To highlight the motivation and characteristics of our proposed dataset, we provide a detailed comparative analysis against traditional static face image datasets (e.g., CelebA~\cite{liu2015deep}, CelebA-HQ~\cite{karras2017progressive}, and LFW~\cite{huang2008labeled}) from two direct perspectives: image content and image format.

\begin{figure}[htbp]
\centering
\includegraphics[width=\linewidth]{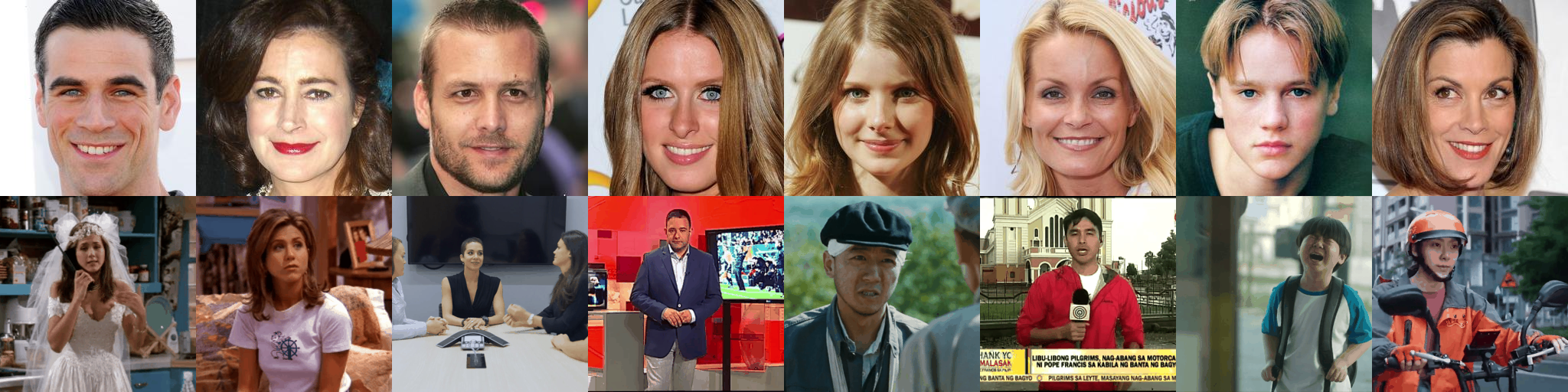}
\caption{Comparison of facial datasets. The top row shows existing face image datasets, and the bottom row shows our constructed GIF face dataset.}\label{fig:s1}
\end{figure}

\textbf{Image Content:} Existing face image datasets often employ a tight cropping strategy. As shown in the first row of Fig.~\ref{fig:s1}, the facial regions in these datasets occupy an overwhelmingly large proportion of the image area (often exceeding 90\%), effectively stripping away environmental context. However, real-world manipulated media frequently involves complex backgrounds. To bridge this gap, our constructed GIFfaces dataset utilizes a context-preserving cropping strategy. While still strictly centered on the face to ensure relevance for deepfake forensics, we expand the bounding box to retain a significantly larger proportion of the surrounding background. As illustrated in the second row of Fig.~\ref{fig:s1}, this multi-scene context makes the spatial distribution of our dataset much closer to real-world scenarios.

\begin{figure}[htbp]
\centering
\includegraphics[width=\linewidth]{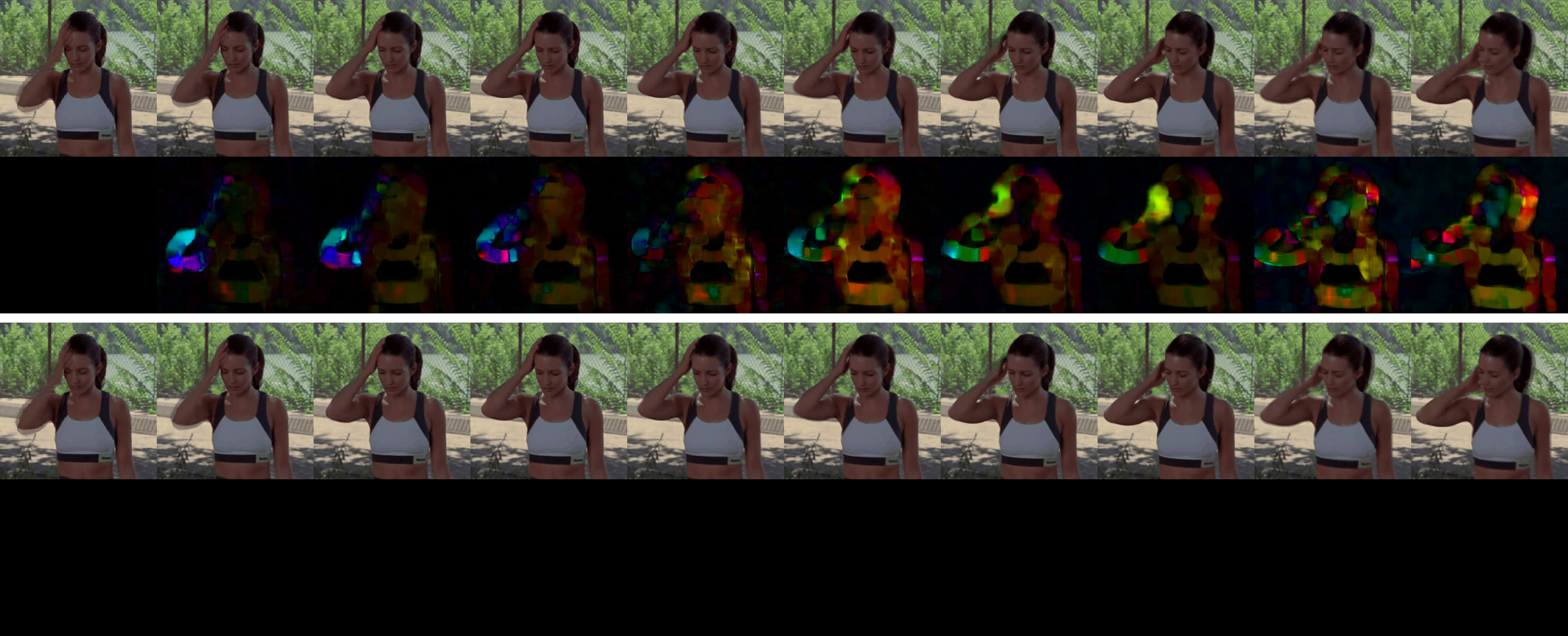}
\caption{Dense optical flow visualization representing natural motion. The top block shows the results of 3D modeling, while the bottom block displays the output of 2D independent frame processing.}\label{fig:s2}
\end{figure}

\begin{figure}[htbp]
\centering
\includegraphics[width=\linewidth]{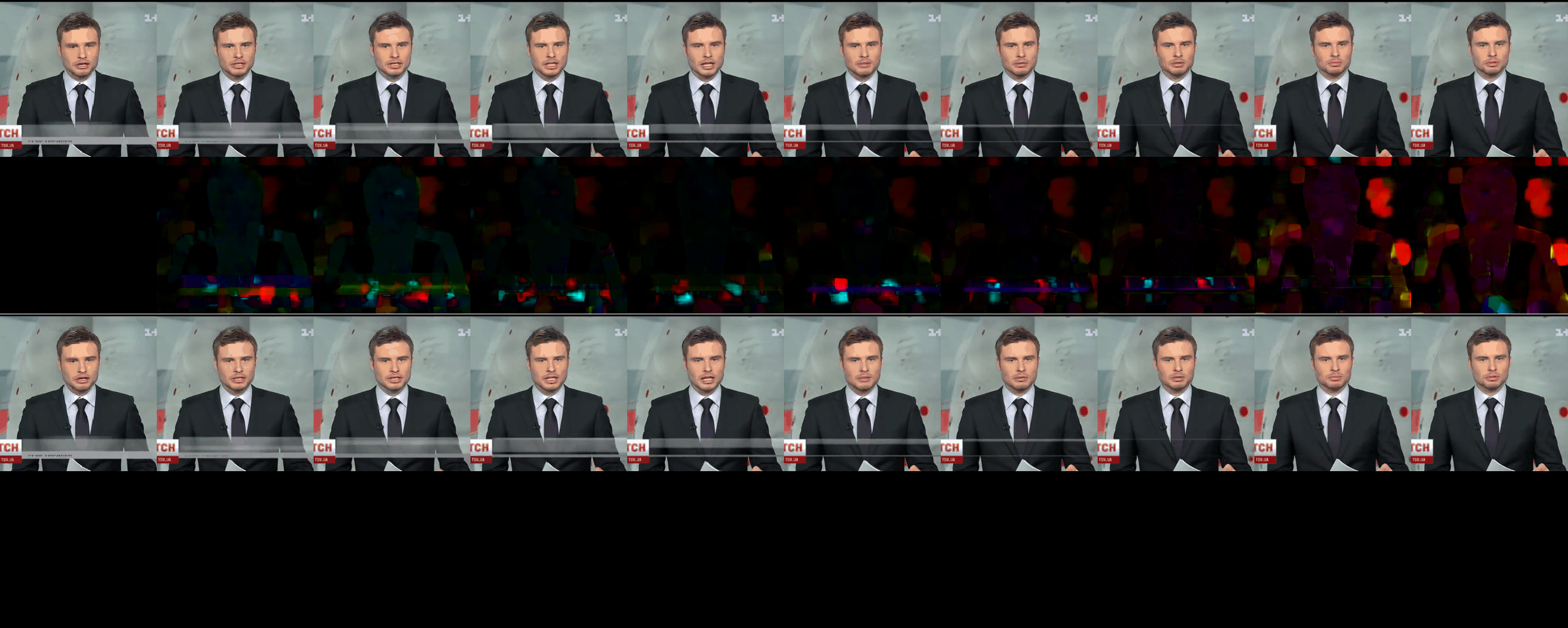}
\caption{Dense optical flow visualization representing facial expressions. The top block shows the results of 3D modeling, while the bottom block displays the output of 2D independent frame processing.}\label{fig:s3}
\end{figure}

\textbf{Image Format:} As discussed in the Introduction, the GIF format conveys richer information by incorporating a continuous temporal flow, simultaneously capturing both spatial details and the evolving temporal dynamics between frames, as shown in Fig.~\ref{fig:s2} and Fig.~\ref{fig:s3}. In these visualizations, the top rows display colored optical flow maps derived from the native GIF sequences, whereas the bottom rows show entirely black frames resulting from treating each GIF frame as an independent static image, which severs the temporal connectivity. This fundamental difference in data format is precisely why many existing proactive forensic methods designed for static images cannot be directly transferred and applied to GIF scenarios.

\begin{figure}[htbp]
\centering
\includegraphics[width=\linewidth]{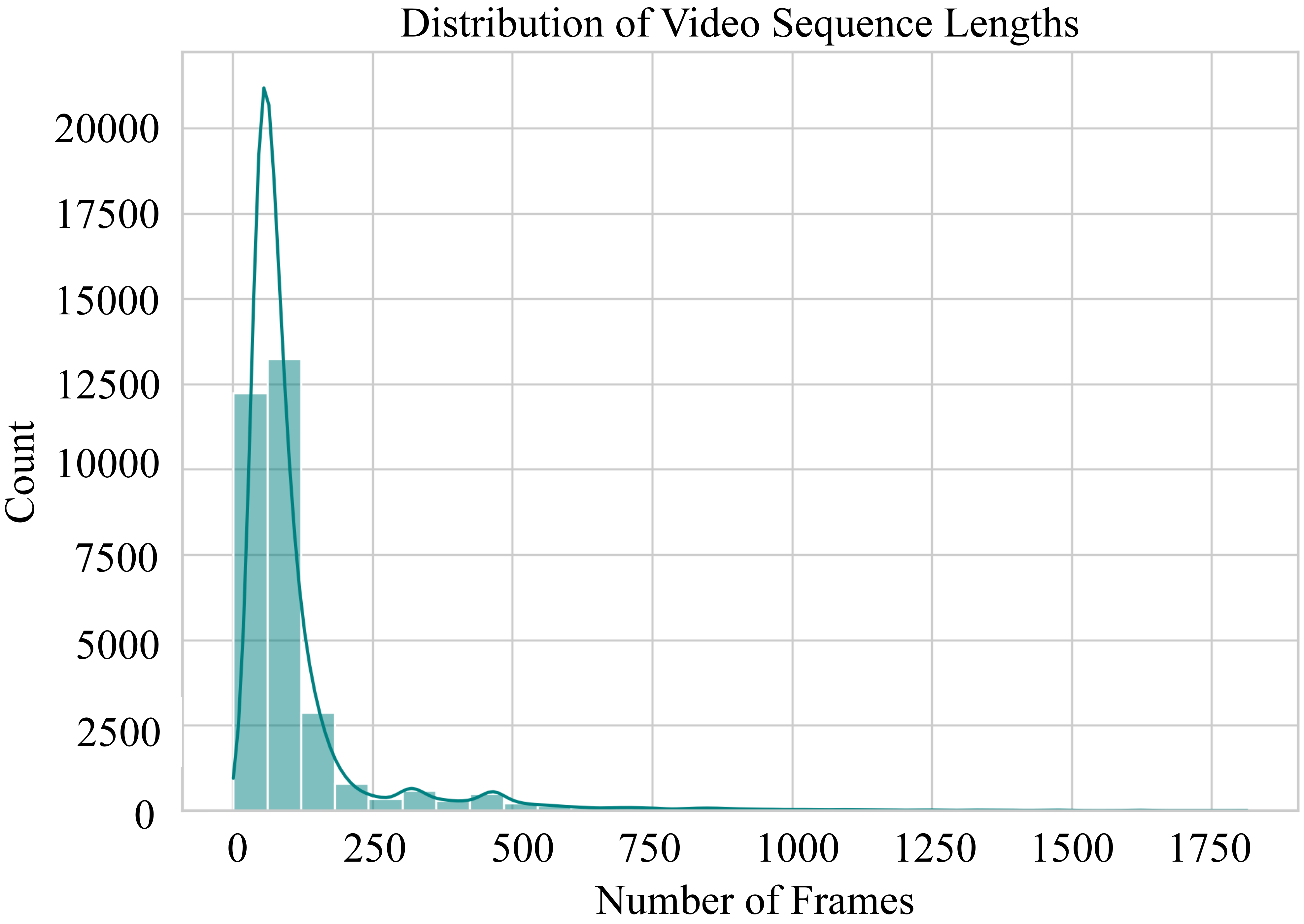}
\caption{Distribution of raw video sequence lengths (frame counts).}\label{fig:s4}
\end{figure}

\begin{figure}[htbp]
\centering
\includegraphics[width=\linewidth]{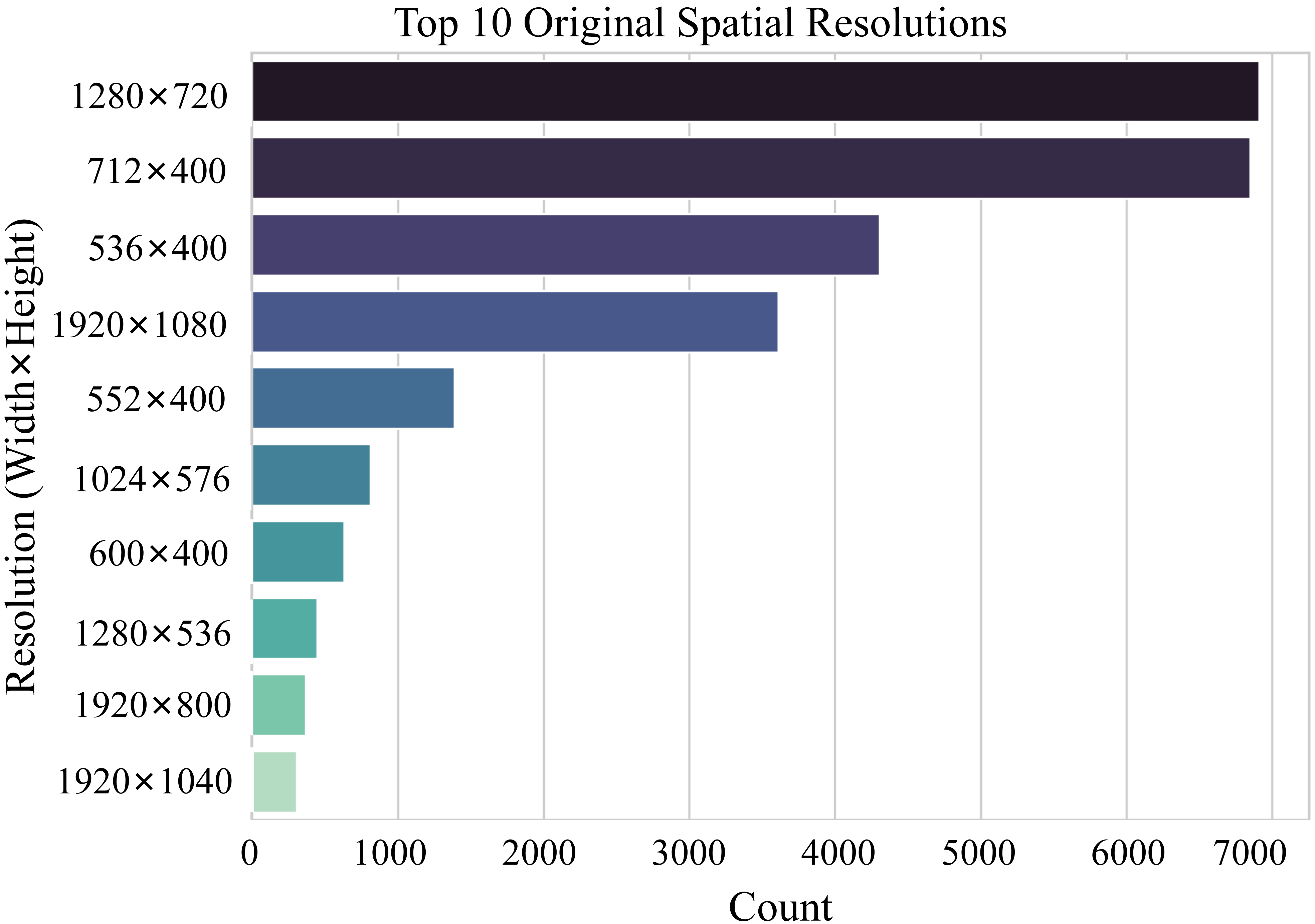}
\caption{Top 10 original spatial resolutions (Width $\times$ Height) of the raw dataset. }\label{fig:s5}\end{figure}

\begin{figure}[htbp]
\centering
\includegraphics[width=\linewidth]{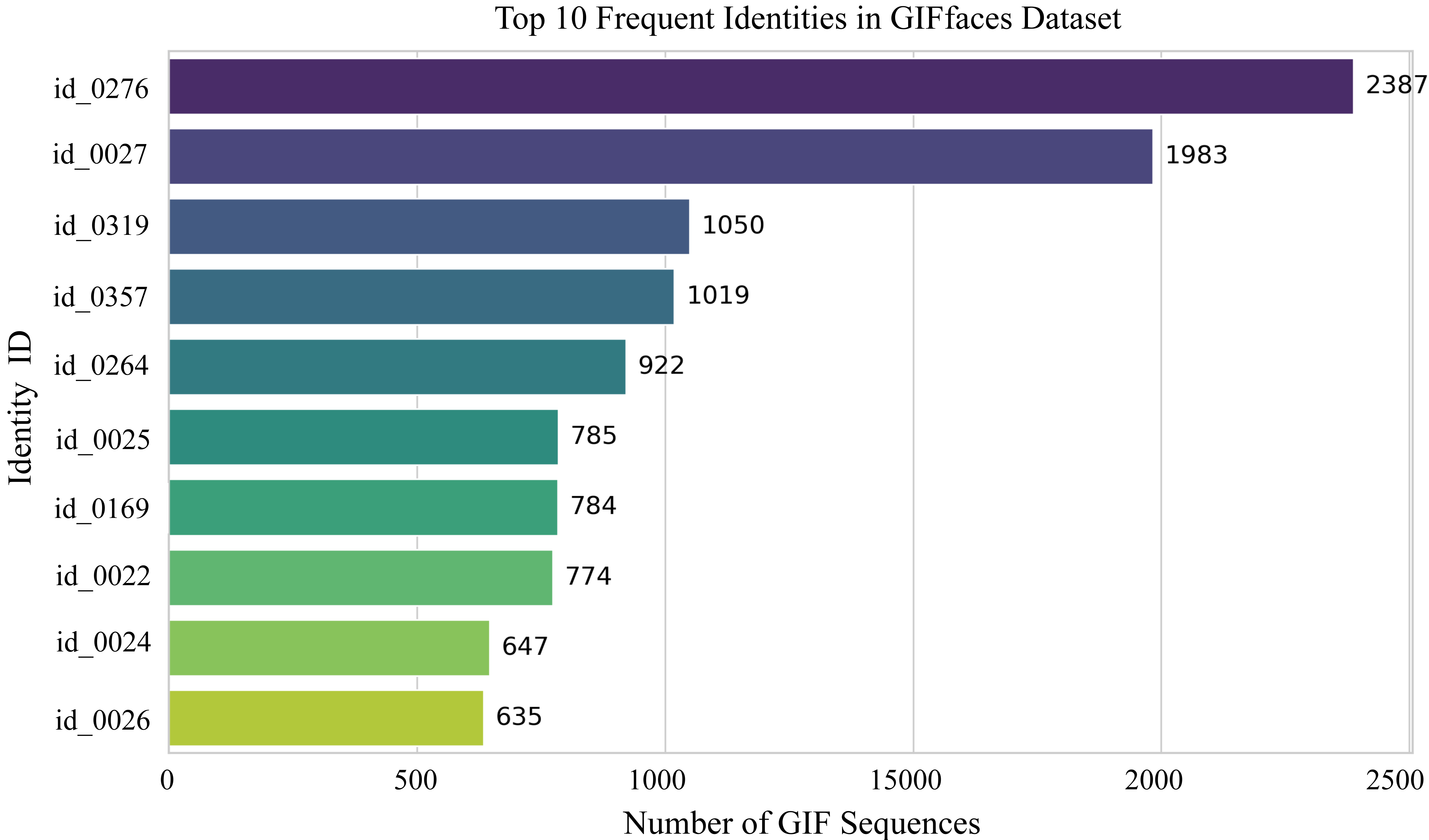}
\caption{Distribution of the top 10 frequent identities in the dataset.}
\label{fig:s6}
\end{figure}

\subsection{Spatiotemporal Variance of the Raw Data}
As detailed in the Methodology section of the main text, our initial raw data pool is aggregated from five highly diverse benchmark datasets alongside unconstrained videos crawled from YouTube. While this comprehensive collection guarantees multi-scenario coverage, it also inevitably introduces extreme variations in video properties and dimensions.

In the temporal dimension, the raw dataset exhibits significant length variations, ranging from short actions to prolonged recordings. This natural variance (partially visualized in Fig.~\ref{fig:s4}) not only creates temporal misalignment but also introduces substantial computational overhead for 3D spatiotemporal modeling. To reduce this computational burden, we extract a fixed 10-frame sequence starting from the first face-detected frame, thereby guaranteeing temporal consistency and training efficiency across all samples.

Similarly, in the spatial dimension, the raw dataset contains multiple resolution formats, including high-definition sizes like $1280 \times 720$ and irregular aspect ratios like $712 \times 400$ (with the top frequency distributions illustrated in Fig.~\ref{fig:s5}). Directly feeding these inconsistent dimensions into a neural network introduces structural and scale biases during feature extraction. To resolve this spatial variance, our processing protocol applies face-centric cropping and resizes all frames to a uniform $256 \times 256$ spatial resolution.

\subsection{Dataset Statistics and Identity Distribution}
To establish a reliable foundation for identity tracking, we  annotated the processed samples, yielding a dataset of over 50,000 GIF sequences distributed across more than 9,000 unique identities. Fig.~\ref{fig:s6} visualizes the sequence counts for the top 10 most frequent identities. This distribution illustrates the scale of our curated data and the availability of multiple temporal variants per subject, providing a diverse data foundation for 3D spatiotemporal modeling.

\section{Efficiency and Parameter Analysis}

We further analyze the efficiency of GIFGuard from three complementary perspectives: trainable parameters, theoretical computational cost, and measured inference speed. This organization separates memory-related model size from actual runtime cost, providing a more complete view of practical deployment.

\subsection{Model Size and Computational Cost}

\begin{table}[htbp]
  \centering
  \caption{Comparison of model efficiency in terms of trainable parameters (M). }\label{tab:s1}
  \resizebox{\linewidth}{!}{%
  \begin{tabular}{lccccc}
    \toprule
    Component & REVmark & GIFMarking & RivaGAN & VideoSeal & GIFGuard (ours) \\
    \midrule
    Encoder   & 4.02    & 0.42       & 0.53    & 103.73      & 22.61         \\
    Decoder   & 12.92   & 0.08       & 1.02    & 97.08      & 406.67        \\
    Adv       & 0.29    & 0.10        & 0       & 0.17    & 0.10         \\
    Total Params & 17.23 & 0.65     & 1.55    & 200.98  & 429.38       \\
    \bottomrule
  \end{tabular}
  }
\end{table}

\begin{table}[htbp]
  \centering
  \caption{Quantitative comparison of theoretical computational cost (GFLOPs). }\label{tab:s2}
  \resizebox{\linewidth}{!}{%
  \begin{tabular}{lccccc}
    \toprule
    Component & REVmark & GIFMarking & RivaGAN & VideoSeal & GIFGuard (ours) \\
    \midrule
    Encoder   & 186.08  & 129.37     & 170.93  & 162.93     & 54.14          \\
    Decoder   & 150.88  & 23.06      & 284.61  & 20.66      & 129.37         \\
    Adv       & 3.85    & 14.45      & 0       & 0.69    & 14.44         \\
    Total FLOPs & 340.81 & 196.01    & 455.53  & 184.28  & 197.95       \\
    \bottomrule
   \end{tabular}%
   }
 \end{table}

\begin{table}[htbp]
  \centering
  \caption{Inference efficiency comparison for deployment.}\label{tab:s3}
  \resizebox{\linewidth}{!}{%
  \begin{tabular}{lcccc}
    \toprule
    & \multicolumn{2}{c}{Encoding} & \multicolumn{2}{c}{Decoding} \\
    \cmidrule(lr){2-3} \cmidrule(lr){4-5}
    Method & Latency (ms) $\downarrow$ & FPS $\uparrow$ & Latency (ms) $\downarrow$ & FPS $\uparrow$ \\
    \midrule
    REVMark & 55.96 & 142.96 & 41.35 & 193.46 \\
    VideoSeal & \textbf{26.12} & \textbf{382.87} & 34.56 & 289.36 \\
    GIFGuard (ours) & 28.15 & 355.20 & \textbf{26.14} & \textbf{382.60} \\
    \bottomrule
  \end{tabular}%
  }
\end{table}

As shown in Table~\ref{tab:s1}, GIFGuard contains 429.38M trainable parameters, with most of them located in the decoder. This mainly comes from the Adaptive Global Projection Head, which preserves rich spatiotemporal features for robust bit recovery instead of compressing them prematurely. Therefore, the large parameter count reflects the memory footprint of the decoder rather than the overall runtime cost alone.

Tables~\ref{tab:s2} and~\ref{tab:s3} further show that model size and deployment efficiency are not strictly coupled. Although GIFGuard has more parameters than several baselines, its total computational cost is 197.95 GFLOPs, which is comparable to GIFMarking and VideoSeal and lower than REVmark and RivaGAN. In measured inference speed, GIFGuard reaches 355.20 FPS for encoding and 382.60 FPS for decoding, matching or exceeding the compared video watermarking baselines. These results indicate that GIFGuard remains practical for real-time deployment, while its main efficiency limitation lies in memory usage. Future work will explore lightweight decoder designs and memory optimization strategies for resource-constrained devices.

\end{document}